\documentclass{ecai} 

\usepackage{latexsym}
\usepackage{amssymb}
\usepackage{amsmath}
\usepackage{amsthm}
\usepackage{booktabs}
\usepackage{enumitem}
\usepackage{graphicx}
\usepackage{color}

\usepackage{wrapfig}
\usepackage{microtype}
\usepackage{graphicx}
\usepackage{subfigure}
\usepackage{booktabs}
\usepackage[hidelinks]{hyperref}

\usepackage{amsmath}
\usepackage{amssymb}
\usepackage{mathtools}
\usepackage{amsthm}
\usepackage[capitalize,noabbrev]{cleveref}
\usepackage[textsize=tiny]{todonotes}

\newcommand{\newchange}{\textcolor{black}}
\newcommand{\baichuanchange}{\textcolor{black}}

\usepackage{amsmath,amsfonts,bm}

\def\eqref#1{equation~\ref{#1}}

\def\1{\bm{1}}

\DeclareMathAlphabet{\mathsfit}{\encodingdefault}{\sfdefault}{m}{sl}
\SetMathAlphabet{\mathsfit}{bold}{\encodingdefault}{\sfdefault}{bx}{n}

\newcommand{\E}{\mathbb{E}}

\usepackage{url}
\usepackage{graphicx}
\usepackage{subcaption}
\usepackage{algorithmic}
\usepackage{algorithm}
\usepackage{array}
\usepackage{natbib}
\usepackage{textcomp}
\usepackage{stfloats}
\usepackage{url}
\usepackage{verbatim}
\usepackage{float} 
\usepackage{multirow}
\usepackage{bm}
\usepackage{url}
\usepackage{xcolor}
\usepackage{booktabs}
\usepackage{pifont}
\usepackage[acronym]{glossaries}
\usepackage{tablefootnote}

\newacronym{EEG}{EEG}{electroencephalogram}
\newacronym{DNNs}{DNNs}{Deep Neural Networks}
\newacronym{ANNs}{ANNs}{Artificial Neural Networks}
\newacronym{ECG}{ECG}{Electrocardiogram}
\newacronym{GPUs}{GPUs}{Graphics Processing Cards}
\newacronym{CPUs}{CPUs}{Central Processing Units}
\newacronym{MACs}{MACs}{Multiply–Accumulate Operations}
\newacronym{TP}{TP}{True Positive}
\newacronym{FP}{FP}{False Positive}
\newacronym{AI}{AI}{Artificial Intelligence}
\newacronym{BP}{BP}{Backpropagation}

\newacronym{IoT}{IoT}{Internet of Things}

\newcommand{\newbold}[1]{\textbf{#1}}
\newcommand{\ourbold}{\ensuremath{\boldsymbol}}
\newcommand{\cameraready}{\textcolor{black}}

\usepackage{orcidlink}

\newcommand{\BibTeX}{B\kern-.05em{\sc i\kern-.025em b}\kern-.08em\TeX}

\glsdisablehyper

\begin{document}


\begin{frontmatter}


\paperid{543} 


\title{LightFF: Lightweight Inference for \\ Forward-Forward Algorithm}


\author[A]{\fnms{Amin}~\snm{Aminifar}\orcidlink{0000-0002-9920-2539}\footnote{Equal contribution. \\ Published at the European Conference on Artificial Intelligence (ECAI), 2024. }}
\author[B]{\fnms{Baichuan}~\snm{Huang}\orcidlink{0000-0003-4010-8545}\footnotemark\thanks{Corresponding Author. Email: baichuan.huang@eit.lth.se.}}
\author[B]{\fnms{Azra}~\snm{Abtahi}\orcidlink{0000-0003-3642-8466}} 
\author[B]{\fnms{Amir}~\snm{Aminifar}\orcidlink{0000-0002-1673-4733}} 

\address[A]{Heidelberg University, Germany}
\address[B]{Lund University, Sweden}


\begin{abstract}
The human brain performs tasks with an outstanding energy efficiency, i.e., with approximately 20 Watts. The state-of-the-art Artificial/Deep Neural Networks (ANN/DNN), on the other hand, have recently been shown to consume massive amounts of energy. The training of these ANNs/DNNs is done almost exclusively based on the back-propagation algorithm, which is known to be biologically implausible. This has led to a new generation of forward-only techniques, including the Forward-Forward algorithm. In this paper, we propose a lightweight inference scheme specifically designed for DNNs trained using the Forward-Forward algorithm. We have evaluated our proposed lightweight inference scheme in the case of the MNIST and CIFAR datasets, as well as two real-world applications, namely, epileptic seizure detection and cardiac arrhythmia classification using wearable technologies, where complexity overheads/energy consumption is a major constraint, and demonstrate its relevance. Our code is available at \href{https://github.com/AminAminifar/LightFF}{https://github.com/AminAminifar/LightFF}.
\end{abstract}

\end{frontmatter}


\setcounter{footnote}{0}

\section{Introduction}
The state-of-the-art \gls{ANNs}/\gls{DNNs} consume massive amounts of energy and pose a threat to the environment~\citep{savazzi2022energy}. A prime example is GPT-3, a Large Language Model (LLM), that consumes over 1000 megawatt-hour for training alone, which is equivalent to a small town's power consumption for a day~\cite{patterson2021carbon}. The training of these \gls{ANNs}/\gls{DNNs} is done almost exclusively based on the back-propagation algorithm, which is known to be biologically implausible \citep{lillicrap2016random}. 
This has led to a wide range of biologically plausible alternatives, e.g., the Forward-Forward algorithm by \cite{hinton2022forward}, focusing on training \gls{DNNs} without resorting to the biologically-implausible back-propagation scheme, to bridge the existing performance--efficiency gap between the \gls{ANNs}/\gls{DNNs} and the cortex.

The majority of the state-of-the-art studies based on the Forward-Forward algorithm have mainly focused on the training of neural networks. However, the inference over already-trained models also consumes a massive amount of energy. Indeed, inference
accounts for approximately $60\%$ of the total machine learning energy used at Google \cite{9810097}. 
This is because the training overheads are per model and often only incurred once, while the inference overheads are per input/usage and, in the long run, the inference overheads may even dominate the overall energy overheads of ANN/DNN, owing to the many-billion-user services that incorporate machine learning.

In this paper, for the first time, we propose a lightweight inference scheme specifically designed for \gls{DNNs} trained using the Forward-Forward algorithm by \cite{hinton2022forward}. The key insight is that the local energy-based techniques, such as the Forward-Forward algorithm, provide a strong intermediate measure to decide whether the local energy or the goodness in the case of the Forward-Forward algorithm is sufficient to make a confident decision, without the need to complete the entire forward pass. Our proposed scheme is inspired by the human nervous system  \cite{saladin2009anatomy}, where the reflexes do not pass directly into the brain, but synapse in the spinal cord, hence without the delay of routing signals through the brain. This is while the complex inputs that require detailed analysis are processed by the brain.

We have evaluated our proposed lightweight inference scheme in the case of the MNIST \citep{lecun1998mnist} and CIFAR \citep{krizhevsky2010cifar} datasets and shown that the inference overheads can be reduced by up to \baichuanchange{10.4} and \baichuanchange{2.2} times, respectively. To demonstrate the relevance of our proposed scheme, we have also evaluated our proposed scheme in the context of two real-world medical applications in the \gls{IoT} domain, namely, epileptic seizure detection using wearable devices \citep{shoeb2009application} and cardiac arrhythmia classification using wearable devices \citep{mark1982annotated}. Wearable technologies, and \gls{IoT} systems in general, are extremely limited in terms of resources, i.e., computing power and energy/battery, and present an excellent application for our proposed lightweight inference to enable real-time and 
long-term monitoring of patients in ambulatory settings.

\section{Lightweight Inference}

The intrinsic characteristic of the Forward-Forward algorithm \citep{hinton2022forward} makes it possible to perform inference without completing a forward pass through all the layers of the network. 
This can be used to save resources when performing inference operations. 
In this section, we present an approach for lightweight inference based on the Forward-Forward algorithm.

Inference for a model trained based on the Forward-Forward algorithm can be performed using two different procedures: In the first procedure, for a test sample, we repeat the forward pass operation for all possible labels and calculate the accumulative goodness. We select the label with the highest accumulative goodness. We refer to this procedure as \textit{multi-pass}, as it requires multiple forward passes. 
In the second procedure, a softmax layer at the head of the network, which is learned in the training phase, is used to infer based on the activity of layers for a test sample with a neutral label. This is referred to as \textit{one-pass}, as it requires a single forward pass. 
Hinton's one-pass inference process is, in essence, similar to that of the PEPITA algorithm proposed by \cite{dellaferrera2022error}, hence our proposed lightweight inference for the one-pass procedure may also be applied to PEPITA, as we show in Section \ref{sec:evaluation}.

\begin{figure*} \centering    
\subfigure[Light Multi-Pass Inference] {
 \label{fig:every_label}     
\includegraphics[width=1\columnwidth]{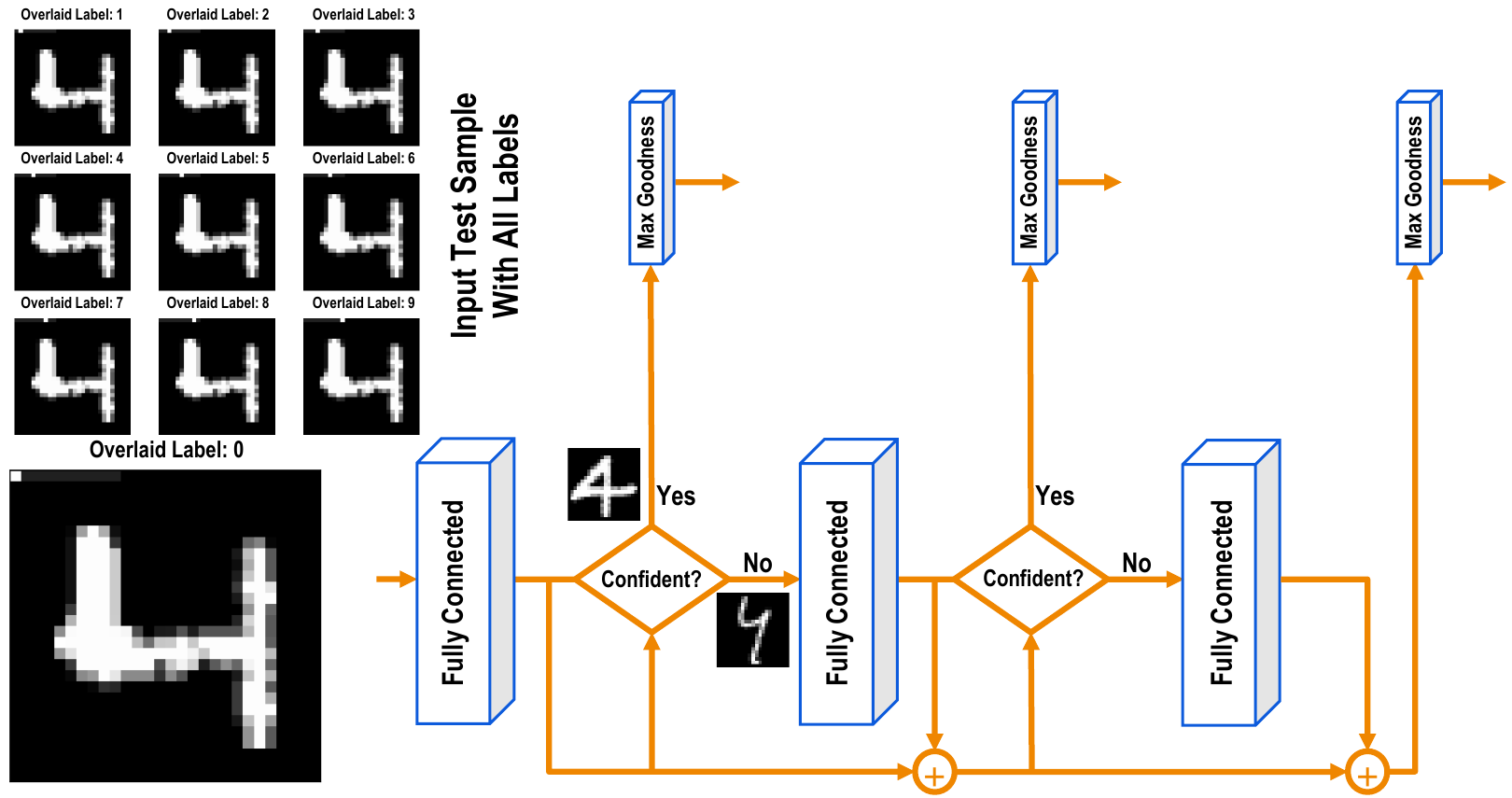}  
}     
\subfigure[Light One-Pass Inference] { 
\label{fig:one_pass}     
\includegraphics[width=1\columnwidth]{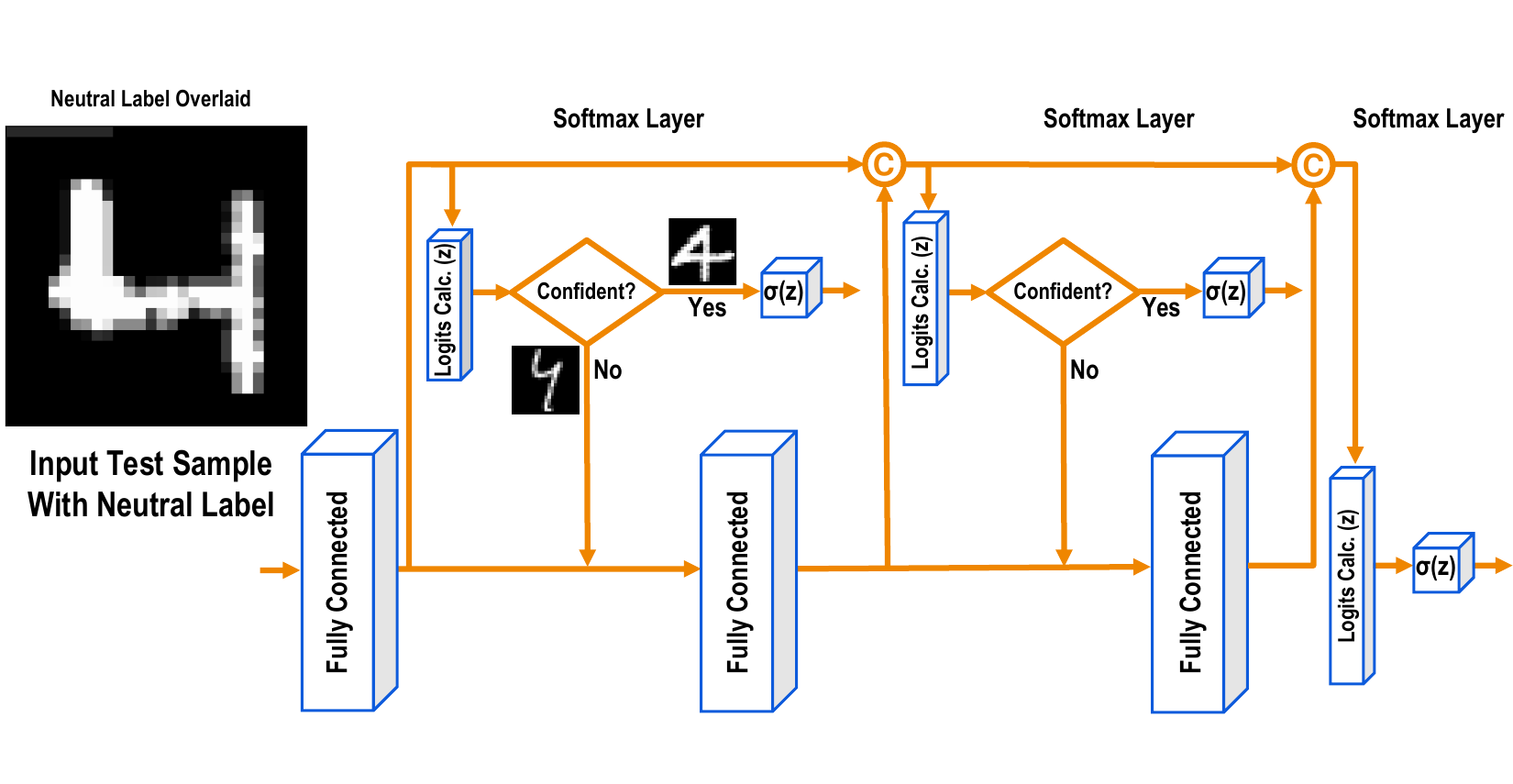}     
}     
\caption{Lightweight Procedures for Inference} \vspace{0.3cm}    
\label{fig:LightweightInferenceProcedures}     
\end{figure*}

\subsection{Inference Based on Multi-Pass Procedure}

In this part, we focus on the multi-pass inference procedure. In our approach, the network's training algorithm and architecture are the same as in \cite{hinton2022forward}. 
For our lightweight inference, however, instead of performing the forward pass through all layers, once the operations for each layer are completed, we inspect the confidence level of the result, and based on that, we decide whether to continue the forward pass. 
The accumulated goodness up to that layer is used to determine the label.

Figure \ref{fig:every_label} shows our lightweight inference scheme. As illustrated in the figure, after each layer, we check the confidence. 
The difficulty of the classification task varies from one test sample to another. 
As shown, the first layer(s) is sufficient for straightforward test samples, while for other samples, more layers may be required.

In the Forward-Forward algorithm, we train the model to have high goodness for the positive samples and low goodness for negative ones. Considering this, we use the magnitude of goodness to decide how confident we are about our inference result. The goodness in one layer is the sum of the squared activities, $\sum_{i=1}^{n} {a_i}^2$, where $a_i$ is the output of $i$th neuron and $n$ is the number of neurons in that layer. For the confidence, we only need to add one single neuron with a sigmoid activation function in each layer whose weights are the activities of that layer and the layers before, and its bias is our confidence threshold. 

\begin{figure*}[ht!]
     \centering
\includegraphics[width=1.0\textwidth]{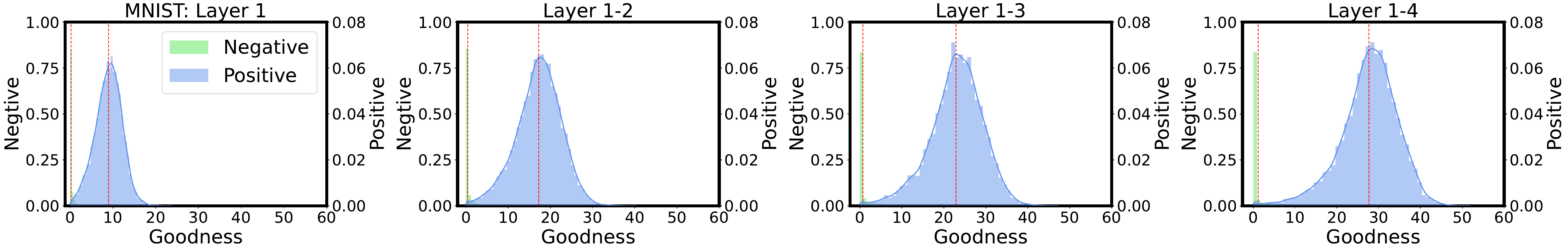}
     \caption{The distribution of MNIST validation data (Green: Negative data; Blue: Positive data) and the corresponding mean values (red vertical lines) in the Forward-Forward algorithm. The distance between the mean values of the negative and positive samples/distributions increases as we consider more layers. }
     \label{fig:mnist_threshold_level} \vspace{0.3cm}
\end{figure*}

The confidence threshold for each layer (i.e., the bias of the neuron introduced in each layer) is learned based on the validation set. 
Considering the fully connected network that is already trained by the Forward-Forward algorithm, the input of validation set $\ourbold{X}^{(l)}$ and the constructed binary label $\ourbold{\bar{y}}^{(l)}$ for layer $l$ are leveraged for training the confidence threshold (i.e., bias) for layer $l$. We define the ground-truth label $\ourbold{\bar{y}}^{(l)}_i=1$ if the input sample $\ourbold{X}^{(l)}_i$ is correctly classified by the layer and $\ourbold{\bar{y}}^{(l)}_i=0$ otherwise. 
For the confidence threshold $b^{(l)}$ of hidden layer $l$, one single neuron, denoted as $\sigma(\ourbold{w}^{(l)}\cdot \ourbold{a}^{(l)}+b^{(l)} )$, is connected to the corresponding activations $\ourbold{a}^{(l)}$, where $\ourbold{w}^{(l)}$ is the weight between hidden layers $l$ and the single neuron. $\sigma$ is the sigmoid function for the single neuron connected to hidden layers $l$. We set $\ourbold{w}^{(l)}=\ourbold{a}^{(l)}$ to capture the goodness value, hence only the value of $b^{(l)}$ needs to be trained. 
We consider the Binary Cross-Entropy loss function, with $\ourbold{\bar{y}}^{(l)}_i$ as the label for the $i$th sample. Alternatively, we can calculate and set the bias values based on the mean and standard deviation of the goodness for the validation set. As shown in Fig. \ref{fig:mnist_threshold_level}, the distance between the mean values of the negative and positive samples/distributions increases as we consider more layers.

Next, let us discuss the computational overhead for utilizing one layer of a Forward-Forward network that directly impacts the energy consumption of the platform on which the inference is performed. Suppose we pass a test sample through a fully connected hidden layer of Rectified Linear Units (ReLUs) with $n$ neurons. We show the required computations for calculating the activity vector and the goodness for layer $i$ with $C_i^a$ and $C_i^g$, respectively. 
The overall computation required for a forward pass on layer $i$ is $C_i^a+C_i^g$. 
In this case, if our network has $N$ hidden layers, the computation required for calculating the goodness for all layers of the network would be $\sum_{i=1}^{N} (C_i^a+C_i^g)$. In the case of inference based on the multi-pass procedure, if the data has $M$ class labels, then, for each test sample, the computational overhead would be $C_{MP}=M \cdot \sum_{i=1}^{N} (C_i^a+C_i^g)$.

In our lightweight inference scheme, we make a trade-off between classification accuracy and computational efficiency (that leads to energy efficiency). As discussed, we define a confidence threshold for every layer determined by the amplitude of accumulated goodness up to that layer based on a validation dataset. The objective would be to increase the classification performance, which could require using more hidden layers, while decreasing the number of layers used for choosing a label, for the sake of computational efficiency.

We show the computation required for the calculation of goodness for one test sample in our lightweight multi-pass inference by $C_{Light-MP}$. Using our lightweight inference scheme, if we present the probability of completing the inference operation at layer $i$ by $p_i$, then we can show the expected computational overhead for the calculation of goodness for one test sample as follows:
\noindent
\begin{align}\label{eq:cost_mp}
\E (C_{Light-MP}) &= M \cdot \sum_{i=1}^{N} \left(\sum_{j=1}^{i} (C_j^a+C_j^g)\right) \cdot p_i \\
&=M\cdot \sum_{j=1}^{N} (C_j^a+C_j^g) \cdot \left(1-\sum_{i=1}^{j-1} p_i\right), \nonumber
\end{align}
where $\sum_{i=1}^{N}  p_i=1$.\footnote{For $j=1$, we consider  $\sum_{i=1}^{j-1}  p_i=0$.} 
Therefore, we have $(1-\sum_{i=1}^{j-1}  p_i)\le1$.
As a result, in the worst-case scenario, i.e., when for $i < N$, $p_i=0$, and for $i=N$, $p_i=1$, the computational overhead is the same as the inference procedure in \cite{hinton2022forward}, i.e., $C_{MP}=M \cdot \sum_{i=1}^{N} (C_i^a+C_i^g)$.

\subsection{Inference Based on One-Pass Procedure}

In this part, we focus on the one-pass inference procedure. 
In general, one-pass inference is more efficient compared to the previous procedure, as instead of passing the test sample with all labels through the network, a test sample is passed through the network once with a neutral label. 
For our lightweight one-pass inference procedure, instead of only one softmax layer at the head of the network, we train one softmax layer for each hidden layer. 
The input of each softmax layer is the concatenated activity vectors of all the previous hidden layers. The softmax layers are trained based on train samples with neutral labels (and after all the hidden layers are trained based on positive and negative training samples).

Figure \ref{fig:one_pass} shows our lightweight one-pass inference scheme. 
The typical configuration of a softmax layer includes calculating the logits and applying the softmax activation function. 
As the size of the input to the softmax layer increases in each layer by concatenation of activity vectors, the size of the Logit Calculation block correspondingly increases at each layer. 
These details are visually represented in the figure. 
After each layer, the activity is passed to its respective softmax layer.  
We inspect the confidence based on the calculated logits for that softmax layer. 
As illustrated in the figure, for test samples with less complexity, the first layer(s) is sufficient, but for more complex samples, we continue the forward pass. 

To capture confidence, we incorporate one neuron with a sigmoid activation function in each softmax layer, which is trained based on the validation dataset. 
The input to this neuron is the maximum logit value for that softmax layer. 
\newchange{Similarly, considering the fully connected network that is already trained by the Forward-Forward algorithm, the input of validation set $\ourbold{X}^{(l)}$ and the constructed binary label $\ourbold{\bar{y}}^{(l)}$ are leveraged 
for training and capturing the confidence for layer $l$. We define the ground-truth label $\ourbold{\bar{y}}^{(l)}_i=1$ if the input sample $\ourbold{X}^{(l)}_i$ is correctly classified by the layer and $\ourbold{\bar{y}}^{(l)}_i=0$ otherwise. 
As discussed, the input to the neuron $\sigma(\ourbold{w}^{(l)}\cdot\ourbold{z}^{(l)}+b^{(l)})$ is the logit value $\ourbold{z}^{(l)}$, where $\ourbold{w}^{(l)}$ and $b^{(l)}$ are the weight and the bias of the single neuron to capture confidence. We consider the Binary Cross-Entropy loss function, with $\ourbold{\bar{y}}^{(l)}_i$ as the label for the $i$th sample.}
Alternatively, we can consider the maximum logit value and calculate and set the bias values based on the mean and standard deviation of the input values (of confidence neurons) for the validation set.

The computational overhead for the one-pass procedure is slightly different from the previous part. First, we do not pass the test sample $M$ times through the network. Second, instead of calculating the goodness of the network's layers for inference, we pass the output of the layers (activity vector) to the softmax layer. However, calculating the activity of each layer is similar to the previous procedure (shown as $C_i^a$ for layer $i$). 
We show the required computations for the $i$th softmax layer with $C_i^s$. 
The overall computation required for a forward pass on layer $i$ is $C_i^a+C_i^s$.
Assuming our network has $N$ hidden layers, making an inference operation by the approach in \cite{hinton2022forward}, requires $C_{OP}=\sum_{i=1}^{N}{C_i^a} + C_N^s$ computational overhead.

We show the computation required for one lightweight one-pass inference operation by $C_{Light-OP}$. Using our lightweight inference scheme, if we present the probability of completing the inference operation at layer $i$ by $p_i$, then we can show the expected computational overhead for making an inference for one test sample as follows:
\noindent
\begin{align}\label{eq:cost_op}
\E (C_{Light-OP}) &=  \sum_{i=1}^{N}\left(\sum_{j=1}^{i} (C_j^a+C_j^s)\right) \cdot p_i\\
&=\sum_{j=1}^{N} (C_j^a+C_j^s) \cdot \left(1-\sum_{i=1}^{j-1} p_i\right), \nonumber
\end{align}
where $\sum_{i=1}^{N}  p_i=1$. 
In the worst-case scenario, i.e., when $i < N$, $p_i$ is equal to zero, and for $i=N$, $p_i$ is equal to one, the computational overhead is $\sum_{i=1}^{N} (C_i^a+C_i^s)$, which is slightly ($\sum_{i=1}^{N-1} C_i^s$) worse than the approach in \cite{hinton2022forward}. 
However, as in most datasets, we have samples with variant complexities, the worst-case scenario is not very likely. 
The first few layers of the model are sufficient for the majority of samples. We experimentally demonstrate this in Section \ref{sec:evaluation}. On the other hand, $C_i^a$ is usually significantly larger than $C_i^s$ because, in most cases, the number of neurons $n$ in each layer is much larger than the number of classes $M$. 
Therefore, even if a small proportion of samples use fewer layers, $C_i^s$ overheads can be compensated.

\section{Evaluation}\label{sec:evaluation}

\subsection{Experimental Setup}

\subsubsection{Datasets}

To evaluate our lightweight inference scheme, we consider the MNIST dataset of handwritten digits \citep{lecun1998mnist} and the CIFAR-10 dataset of object recognition \citep{krizhevsky2010cifar}. In addition, we also evaluate our proposed scheme in the context of two real-world medical applications, namely, epilepsy monitoring and seizure detection using wearable technologies based on the CHB-MIT Scalp \gls{EEG} Dataset \citep{shoeb2009application} and cardiac arrhythmia classification using wearable technologies based on the MIT-BIH Arrhythmia \gls{ECG} Dataset \citep{mark1982annotated}, where complexity overhead/energy consumption is a major constraint to ensure real-time and long-term monitoring in ambulatory settings.

\textbf{MNIST \citep{lecun1998mnist}:} The MNIST dataset contains handwritten digits, which are $28\times28$ grayscale images in 10 different classes (one for each of the 10 digits). We used 50,000 images for training, 10,000 images for validation, and 10,000 images for testing in the inference.

\textbf{CIFAR-10 \citep{krizhevsky2010cifar}:} The CIFAR-10 dataset is a Computer Vision dataset for object recognition, which consists of $32\times32$ color images in 10 different classes. We used 45,000 images for training, 10,000 images for validation, and 5,000 images for testing in the inference.

\textbf{CHB-MIT Scalp \gls{EEG} Dataset \citep{shoeb2009application}:} This dataset contains \gls{EEG} recordings,  collected from 22 patients with epilepsy (5 males and 17 females). The recordings are grouped into 23 cases and are divided into seizure and non-seizure classes. To be consistent with wearable IoT devices for real-time seizure monitoring \citep{sopic2018glass}, only the data corresponding to two channels, i.e., T7F7 and T8F8, are considered for the classification. The data corresponding to patients 6, 14, and 16 is eliminated due to their very short-lasting seizures. We use 70\%, 15\%, and 15\% of the dataset for training, validation, and test process, respectively.

\textbf{MIT-BIH Arrhythmia Dataset \citep{mark1982annotated}:} This dataset encompasses \gls{ECG} recordings from 47 different patients with cardiovascular problems.   Five different types of arrhythmias are categorized by beat annotations \citep{arlington1998testing}. We use the pre-processed data provided in \cite{kachuee2018ecg}. We consider 70\% of data for training, 15\% for validation, and 15\% to test the model.
 
\subsubsection{Baselines}
As discussed previously, we apply our proposed lightweight inference in the context of three state-of-the-art techniques, namely:
\begin{itemize}
    \item Forward-Forward, Multi-Pass (MP), \citep{hinton2022forward}: This is the implementation of the original proposal in \cite{hinton2022forward}, where the inference process requires several (as many as the number of classes) forward passes, hence referred to as Multi-Pass (MP).\footnote{https://github.com/loeweX/Forward-Forward}
    \item Forward-Forward, One-Pass (OP), \citep{hinton2022forward}: This is the implementation of the efficient inference proposal in \cite{hinton2022forward}, where the inference process requires only one forward pass, hence referred to as One-Pass (OP), which is built on top of the MP implementation.
    \item PEPITA (PT), \citep{dellaferrera2022error}: This is the implementation of the algorithm proposed in \cite{dellaferrera2022error}, which is referred to as PEPITA (PT), where the inference process requires one forward pass. The PEPITA implementation by \cite{dellaferrera2022error}\footnote{https://github.com/GiorgiaD/PEPITA} only supports up to 3 hidden layers \citep{pau2023suitability}. 
\end{itemize}

\begin{table*}[ht]
\caption{Error (\%) Evaluation of Our Lightweight Inference Scheme}
\begin{center}
\scalebox{0.96}{
\begin{tabular}{@{}cc|cc|cc|cc@{}}
\toprule[2pt]
\multirow{3}{*}{Dataset} & \multirow{3}{*}{Layers} & \multicolumn{2}{c|}{Forward-Forward [MP]} & \multicolumn{2}{c|}{Forward-Forward [OP]} & \multicolumn{2}{c}{PEPITA [PT]}        \\
                         &                         & \multicolumn{2}{c|}{\citep{hinton2022forward}}          & \multicolumn{2}{c|}{\citep{hinton2022forward}}        & \multicolumn{2}{c}{\citep{dellaferrera2022error}}              \\ \cline{3-8} 
                         &                         & MP          & Light-MP          & OP         & Light-OP         & PT & Light-PT \\    
\midrule[1pt]
        & 2      &   1.45\text{\tiny$\pm$0.05}   &  \newbold{1.43\text{\tiny$\pm$0.04}}    &  1.45\text{\tiny$\pm$0.05}    & \newbold{1.23\text{\tiny$\pm$0.03}}  & \cameraready{2.19\text{\tiny$\pm$0.04}} & \cameraready{\newbold{1.90\text{\tiny$\pm$0.07}}}\\
        & 3      &    \newbold{1.40\text{\tiny$\pm$0.01}}     &    \newbold{1.40\text{\tiny$\pm$0.02}}   &  1.45\text{\tiny$\pm$0.09}   &  \newbold{1.06\text{\tiny$\pm$0.02}}  & 4.98\text{\tiny$\pm$0.28}&\newbold{4.85\text{\tiny$\pm$0.09}}\\
MNIST   & 4      &   \newbold{1.51\text{\tiny$\pm$0.01}}     &    \newbold{1.51\text{\tiny$\pm$0.02}}   &  1.53\text{\tiny$\pm$0.10} & \newbold{1.02\text{\tiny$\pm$0.07}} & - &-\\
\citep{lecun1998mnist}       & 5 & 1.60\text{\tiny$\pm$0.05} & \newbold{1.53\text{\tiny$\pm$0.01}} & 1.57\text{\tiny$\pm$0.03} & \newbold{1.09\text{\tiny$\pm$0.09}} & - &-\\ 
        & 6      &   1.62\text{\tiny$\pm$0.01}      &  \newbold{1.55\text{\tiny$\pm$0.02}}     &  1.57\text{\tiny$\pm$0.05}     & \newbold{0.93\text{\tiny$\pm$0.04}}  & - &-\\ \hline
        & 2      &   50.99\text{\tiny$\pm$0.09}   &  \newbold{46.17\text{\tiny$\pm$0.27}}    &   47.42\text{\tiny$\pm$0.26}      & \newbold{46.56\text{\tiny$\pm$0.25}}  & \cameraready{49.52\text{\tiny$\pm$0.36}} & \cameraready{\newbold{48.94\text{\tiny$\pm$0.30}}} \\
        & 3      &   50.59\text{\tiny$\pm$0.09}     &   \newbold{46.12\text{\tiny$\pm$0.36}}    &   47.42\text{\tiny$\pm$0.12}     & \newbold{46.41\text{\tiny$\pm$0.32}} & \cameraready{49.30\text{\tiny$\pm$0.29}} & \cameraready{\newbold{48.54\text{\tiny$\pm$0.25}}}\\
CIFAR-10 & 4      &   50.65\text{\tiny$\pm$0.11}    &  \newbold{46.05\text{\tiny$\pm$0.30}}     & 47.78\text{\tiny$\pm$0.37} & \newbold{46.25\text{\tiny$\pm$0.37}} & - &-\\
\citep{krizhevsky2010cifar}  & 5 & 50.37\text{\tiny$\pm$0.46} & \newbold{45.43\text{\tiny$\pm$0.03}} & 48.01\text{\tiny$\pm$0.17} & \newbold{46.30\text{\tiny$\pm$0.23}} & - &-\\
        & 6      & 50.75\text{\tiny$\pm$0.28}  &  \newbold{45.83\text{\tiny$\pm$0.34}}    &   47.67\text{\tiny$\pm$0.23}    & \newbold{46.06\text{\tiny$\pm$0.40}} & - &-\\ \hline
        & 2      &  37.54\text{\tiny$\pm$0.53}   & \newbold{36.22\text{\tiny$\pm$0.40}}    &   36.79\text{\tiny$\pm$0.26}     & \newbold{31.38\text{\tiny$\pm$0.54}}  & 36.08\text{\tiny$\pm$1.44} & \newbold{31.91\text{\tiny$\pm$1.60}} \\
        & 3      &  37.54\text{\tiny$\pm$0.55}    &  \newbold{34.65\text{\tiny$\pm$0.93}}     &  36.22\text{\tiny$\pm$0.55}     & \newbold{28.86\text{\tiny$\pm$0.80}}  & 34.98\text{\tiny$\pm$1.88} & \newbold{31.20\text{\tiny$\pm$0.77}}\\
CHB-MIT  & 4      &  39.62\text{\tiny$\pm$0.55}    &  \newbold{34.84\text{\tiny$\pm$1.45}}     & 37.98\text{\tiny$\pm$0.98} & \newbold{28.23\text{\tiny$\pm$0.08}} & - &-\\
\citep{shoeb2009application} & 5 & 38.74\text{\tiny$\pm$0.18} & \newbold{32.29\text{\tiny$\pm$0.82}} & 37.29\text{\tiny$\pm$1.64} & \newbold{25.15\text{\tiny$\pm$0.17}} & - &-\\
        & 6      &  39.18\text{\tiny$\pm$0.35}  &  \newbold{32.07\text{\tiny$\pm$0.53}}   & 37.29\text{\tiny$\pm$0.72}   & \newbold{24.02\text{\tiny$\pm$0.23}} & - &-\\ \hline
        & 2      &   10.82\text{\tiny$\pm$0.63}  &  \newbold{10.06\text{\tiny$\pm$0.44}}  & 10.44\text{\tiny$\pm$0.08}     &  \newbold{10.13\text{\tiny$\pm$0.19}} & 16.05\text{\tiny$\pm$0.41} & \newbold{13.98\text{\tiny$\pm$0.50}} \\
        & 3      & 10.29\text{\tiny$\pm$0.16}   & \newbold{9.68\text{\tiny$\pm$0.34}}  & 10.82\text{\tiny$\pm$0.66}   & \newbold{10.68\text{\tiny$\pm$0.29}} & 15.19\text{\tiny$\pm$0.40} & \newbold{14.31\text{\tiny$\pm$0.21}}\\
MIT-BIH  & 4      &  10.74\text{\tiny$\pm$0.97}    & \newbold{10.07\text{\tiny$\pm$0.21}}   & 10.86\text{\tiny$\pm$0.13} & \newbold{10.54\text{\tiny$\pm$0.14}} & - &-\\
\citep{mark1982annotated}    & 5 & 10.87\text{\tiny$\pm$0.18} & \newbold{9.84\text{\tiny$\pm$0.25}}  & 10.93\text{\tiny$\pm$0.61} & \newbold{10.11\text{\tiny$\pm$0.21}} & - &-\\
        & 6      &  10.66\text{\tiny$\pm$0.04}  & \newbold{9.54\text{\tiny$\pm$0.41}} &  10.59\text{\tiny$\pm$0.39}    & \newbold{9.99\text{\tiny$\pm$0.12}}  & - &- \\ 
        \bottomrule[2pt]
 \end{tabular}}
\label{tab:goodness_1}
\end{center}
\end{table*}

\begin{table*}[ht]
\caption{Complexity (MACs) Improvement of Our Lightweight Inference Scheme (Mean Layers is shown in parentheses)}\vspace{0.2cm}
\centering
\scalebox{0.91}{
\begin{tabular}{@{}cc|cc|cc|cc@{}}
\toprule[2pt]
\multirow{3}{*}{Dataset} & \multirow{3}{*}{Layers} & \multicolumn{2}{c|}{Forward-Forward [MP]} & \multicolumn{2}{c|}{Forward-Forward [OP]} & \multicolumn{2}{c}{PEPITA [PT]}        \\
                         &                         & \multicolumn{2}{c|}{\citep{hinton2022forward}}          & \multicolumn{2}{c|}{\citep{hinton2022forward}}        & \multicolumn{2}{c}{\citep{dellaferrera2022error}}              \\ \cline{3-8} 
                         &                         & MP          & Light-MP          & OP         & Light-OP         & PT & Light-PT \\ 

\midrule[1pt]
        & 2      & 5.31M & \textbf{2.18M} (1.18)   & 5.35M &  \textbf{1.57M} (1.01)  & 1.76M & \cameraready{\textbf{0.95M}} (\cameraready{1.19})\\ 
        & 3      & 9.12M &  \textbf{2.56M} (1.28)  & 9.18M & \textbf{1.61M} (1.03) & 2.76M& \textbf{1.14M} (1.37)\\ 
MNIST   & 4      & 12.94M & \textbf{2.95M} (1.38) & 13.01M & \textbf{1.65M} (1.04) & - &- \\
\citep{lecun1998mnist}       & 5 & 16.75M & \textbf{3.29M} (1.47) & 16.85M & \textbf{1.69M} (1.05) & - &-\\
        & 6      & 20.57M & \textbf{3.67M} (1.57)  & 20.68M & \textbf{1.73M} (1.06) & - &- \\ \hline
        & 2      & 9.67M &  \textbf{7.76M} (1.49)      & 9.71M & \textbf{7.47M} (1.42)  & 4.00M & \cameraready{\textbf{3.25M}} (\cameraready{1.25})\\ 
        & 3     & 13.49M &  \textbf{9.10M} (1.85)    & 13.54M &  \textbf{8.90M} (1.79)  & 5.00M & \cameraready{\textbf{3.43M}} (\cameraready{1.43}) \\ 
CIFAR-10 & 4    & 17.30M & \textbf{10.31M} (2.16) & 17.38M & \textbf{10.34M} (2.16) & - &- \\
\citep{krizhevsky2010cifar}  & 5 & 21.12M & \textbf{11.55M} (2.49) & 21.21M & \textbf{11.80M} (2.54)& - &- \\
        & 6      &  24.93M &  \textbf{12.62M} (2.72)  & 25.05M & \textbf{13.13M} (2.89) & - &- \\ \hline
        & 2      & 5.77M &  \textbf{4.54M} (1.67)    & 5.80M & \textbf{3.26M} (1.33) & 2.00M& \textbf{1.17M} (1.16) \\ 
        & 3      & 9.58M &  \textbf{6.26M} (2.12)   & 9.64M & \textbf{4.52M} (1.66)  & 3.00M & \textbf{1.25M} (1.25)  \\ 
CHB-MIT  & 4      & 13.39M & \textbf{8.39M} (2.68) & 13.47M & \textbf{5.62M} (1.95) & - &- \\
\citep{shoeb2009application} & 5 & 17.21M & \textbf{10.19M} (3.16) & 17.31M & \textbf{6.62M} (2.21) & - &- \\
        & 6      & 21.02M &   \textbf{11.02M} (3.38)   & 21.14M & \textbf{7.43M} (2.42) & - &- \\ \hline 
        & 2      & 4.17M &  \textbf{1.07M} (1.18)    & 4.21M & \textbf{0.71M} (1.08)  & 1.18M & \textbf{0.44M} (1.25)\\ 
        & 3      & 7.98M &  \textbf{1.49M} (1.29)   & 8.04M & \textbf{1.02M} (1.16)  & 2.18M &  \textbf{0.73M} (1.54)\\ 
MIT-BIH  & 4      & 11.80M & \textbf{1.99M} (1.43)  & 11.88M & \textbf{1.31M} (1.24) & - &- \\
\citep{mark1982annotated}    & 5 & 15.61M & \textbf{2.01M} (1.43) & 15.71M & \textbf{1.59M} (1.30) & - &- \\
        & 6      &19.43M & \textbf{3.06M} (1.71) & 19.53M & \textbf{1.84M} (1.38) & - &- \\
        \bottomrule[2pt]
\end{tabular}}
\label{tab:onepass_1}
\vspace{0.3cm}
\end{table*}

\begin{table}[ht]
\vspace{0.4cm}
\caption{Execution Time (in millisecond) for Our Scheme Against the Forward-Forward (FF) Algorithm}\vspace{0.2cm}
\centering
\scalebox{0.91}{
\begin{tabular}{@{}cc|cc|cc@{}}
\toprule[2pt]
\multirow{2}{*}{Dataset} &  \multirow{2}{*}{Layers} & \multicolumn{2}{c|}{FF [MP]} & \multicolumn{2}{c}{FF [OP]}        \\ \cline{3-6} 
                         &                         & MP          & Light-MP          & OP         & Light-OP       \\ 

\midrule[1pt]
        & 2      & 6.58 &  \textbf{2.24}  & 0.72 & \textbf{0.30}  \\ 
        & 3      & 14.77 &  \textbf{3.25} & 1.40  &\textbf{0.31} \\ 
MNIST   & 4      & 22.81 & \textbf{3.39} & 2.30 & \textbf{0.32}  \\
      & 5 & 29.70 & \textbf{4.22} & 3.06 & \textbf{0.34} \\
        & 6      & 36.73 & \textbf{5.52} & 3.75 & \textbf{0.36}\\ \hline
        & 2      & 14.85 & \textbf{10.38} & 1.39 & \textbf{1.15}\\ 
        & 3     & 22.32 & \textbf{13.19}   &2.20  &\textbf{1.47}  \\ 
CIFAR-10 & 4    & 29.57 & \textbf{16.38} & 2.98 & \textbf{1.76} \\
  & 5 & 35.83 & \textbf{17.45} & 3.36 & \textbf{2.01}  \\
        & 6      & 42.38  &  \textbf{19.50}   & 4.31  & \textbf{2.24}  \\ \hline
        & 2      & 1.55 & \textbf{1.19}   & 0.77  & \textbf{0.53}  \\ 
        & 3      & 3.42 & \textbf{1.94}  & 1.42 &  \textbf{0.76} \\ 
CHB-MIT  & 4      & 4.73  & \textbf{2.85}  & 2.28 & \textbf{1.01}  \\
 & 5 & 6.02  & \textbf{3.51} & 3.44 & \textbf{1.21}\\
        & 6      & 7.33 & \textbf{3.74}  & 3.72 & \textbf{1.36} \\ \hline 
        & 2      & 2.56 & \textbf{0.88} & 0.58 & \textbf{0.23} \\ 
        & 3      & 6.83 & \textbf{1.36} & 1.18 & \textbf{0.28} \\ 
MIT-BIH  & 4      & 11.26 & \textbf{1.83} & 1.95 & \textbf{0.34} \\
    & 5 & 14.46 & \textbf{1.87} & 2.88 & \textbf{0.41} \\
        & 6  & 18.03 & \textbf{2.90} & 3.49  &\textbf{0.44}  \\
        \bottomrule[2pt]
\end{tabular}}
\label{tab:time}
\vspace{0.3cm}
\end{table}

\subsubsection{Implementation Details}
Our proposed lightweight inference scheme is implemented in PyTorch \citep{paszke2019pytorch}. We have trained all three state-of-the-art algorithms and tested our lightweight inference scheme on the server of $2\times16$-core Intel(R) Xeon(R) Gold 6226R (Skylake) \gls{CPUs} and 1 NVIDIA Tesla T4 \gls{GPUs}. 

For training MP and OP, we consider several scenarios for fully connected networks with a maximum of 6 hidden layers and a maximum of 2,000 neurons for each layer. We also consider the model used by \cite{hinton2022forward} (4 hidden layers and 2000 neurons for each layer) as the {default model} in MP and OP. We use the Stochastic Gradient Descent (SGD) optimizer with a learning rate of $0.001$. We set the batch size to 100 for the MNIST, CIFAR-10, and MIT-BIH datasets. For the CHB-MIT dataset, the personalized model was trained with batch gradient descent.  As for other training parameters, we have considered the values according to the Forward-Forward algorithm. 

For training PT, we consider the model used by \cite{srinivasan2023forward} (3 hidden layers and 1024 neurons for each layer) as the {default model} in PT. We use a momentum optimizer. We set the batch size to 64 for MNIST, CIFAR-10, and MIT-BIH datasets. We set the number of epochs to 100 for all four datasets in MP, OP, and PT. After training, we extract the confidence threshold for each layer based on the validation data.

The datasets in our experiments are balanced, and error, i.e., the total number of incorrectly classified inputs divided by the total number of inputs, is used as the metric of classification performance. We use \gls{MACs} to evaluate the complexity of inference processes. In addition, each test sample exploits the corresponding number of layers used in lightweight inference. We use the ``Mean Layers'' to capture the average/mean number of layers used for all the test samples. 

\subsection{Results}
In this section, we evaluate our lightweight inference scheme in terms of prediction performance/error and computational complexity. 

\subsubsection{Performance Evaluation}

In this section, we investigate the performance of our lightweight inference scheme based on MP, OP, and PT. We consider the networks with the same number of neurons (2000 neurons for MP and OP; 1024 neurons for PT) in each hidden layer. We adjust the number of hidden layers from 2 to 6 for MP and OP. We adjust the number of hidden layers from 2 to 3 for PT because PT only supports up to 3 hidden layers \citep{pau2023suitability}. 

Table \ref{tab:goodness_1} shows the error comparison between these three state-of-the-art techniques and our lightweight inference scheme. For MP, the Light-MP error is on par with the MP error for MNIST, CIFAR-10, CHB-MIT, and MIT-BIH datasets in different layer settings. Similarly, for OP, the Light-OP error is on par with the OP error for all four datasets. \cameraready{Finally, for PT, the Light-PT error is also comparable to the PT error for all four datasets}.\footnote{Note that, as shown in \cite{srinivasan2023forward}, the overall error marginally increases with the number of layers.} 
In summary, our lightweight inference achieves a comparable classification error and even a smaller error in some cases.

\subsubsection{Complexity Evaluation}

In this section, we evaluate the complexity of our lightweight inference scheme based on MP, OP, and PT. First, we refer to the same experimental settings in Table \ref{tab:goodness_1} and report the \gls{MACs} and Mean Layers (shown in parentheses) used by our lightweight inference scheme in Table \ref{tab:onepass_1}. For Light-MP, Light-OP, and Light-PT evaluated on MNIST, CIFAR-10, CHB-MIT, and MIT-BIH datasets, the value of Mean Layers used in our lightweight inference scheme is always significantly lower than the number of the network layers.  Similarly, the \gls{MACs} of Light-MP, Light-OP, and Light-PT are always significantly lower than the \gls{MACs} of MP, OP, and PT, respectively, for all four datasets in different layer settings. However, as the number of network layers increases from 2 to 6 for MP and OP, and from 2 to 3 for PT, the Mean Layers used tend to increase but the relative percentage of layers used decreases. Moreover, we have conducted experiments to show the actual execution time reduction of our lightweight inference scheme, on a MacBook Pro with the Apple M1 pro CPU and 32 GB of RAM. Table \ref{tab:time} presents the improvement of our scheme against the Forward-Forward algorithm. Taking MNIST with 4 layers as an example, the execution time of FF [MP], i.e.,  22.81 ms, is reduced by a factor of $6.7\times$ to 3.39 ms, and the execution time of FF [OP], i.e., 2.30 ms, by a factor of $7.2\times$ to 0.32 ms.

Hence, considering Table \ref{tab:goodness_1}, Table \ref{tab:onepass_1}, and Table \ref{tab:time}, we conclude that our lightweight inference scheme improves computational efficiency by reducing the number of layers used in inference, with a comparable classification error in all cases.

\begin{figure*}[ht]
     \centering
\includegraphics[width=1.0\textwidth]{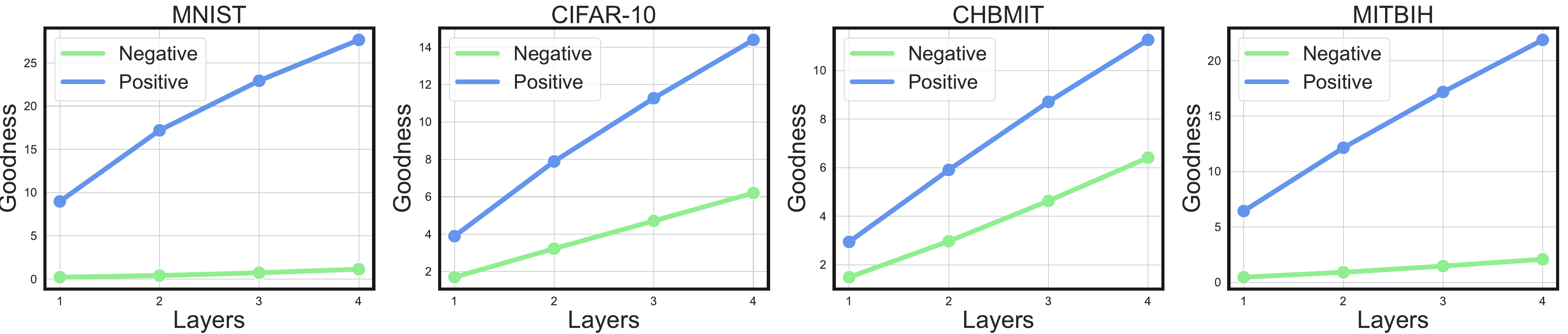}\vspace{0.2cm}
     \caption{The mean values of negative data and positive data in the Forward-Forward algorithm (MP).}\vspace{0.3cm}
     \label{fig:MP_threshold_distance}
\end{figure*}

\begin{figure*}[ht]
     \centering
\includegraphics[width=0.8\textwidth]{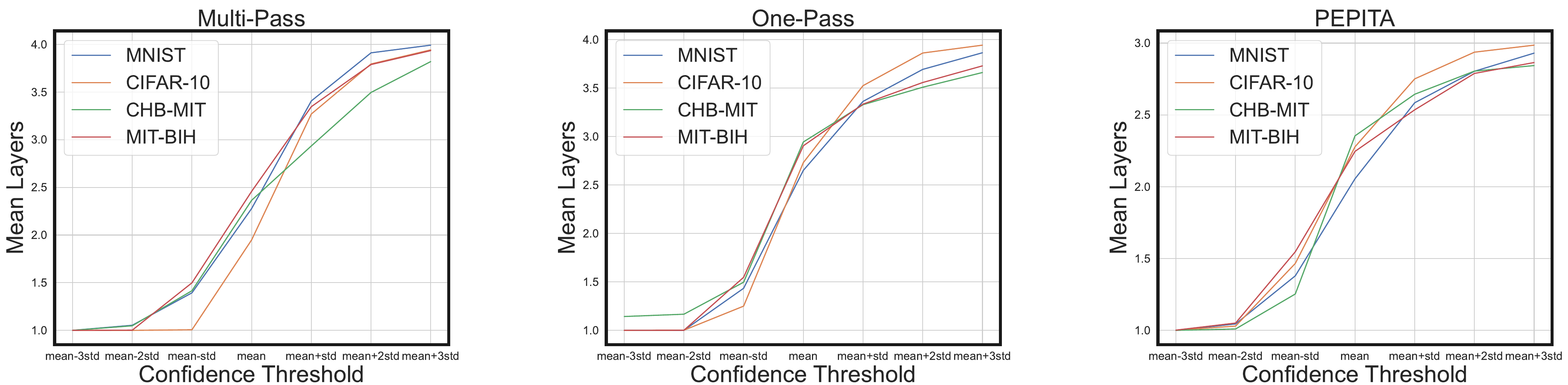}\vspace{0.2cm}
     \caption{\cameraready{The average/mean number of layers used by our lightweight inference schemes versus confidence threshold.}}\vspace{0.3cm}
     \label{fig:different_threshold}
\end{figure*}

\begin{figure*}[ht]
     \centering
\includegraphics[width=0.9\textwidth]{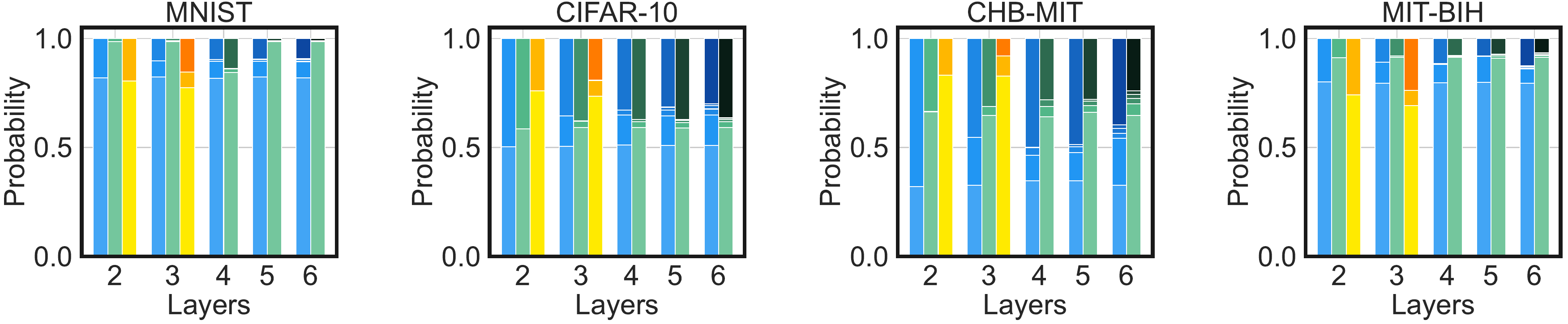}\vspace{0.2cm}
     \caption{\cameraready{Probability of the number of layers used, for 2--6 layer networks, four datasets, and three algorithms (MP, OP, and PT). Blue: MP; Green: OP; Yellow: PT. The lightest color represents 1 layer and the darkest color represents 6 layers.} }
     \vspace{0.3cm}
     \label{fig:different_layer_probability}
\end{figure*}

\subsubsection{Detailed Analysis and Discussion}

To show the relevance of our lightweight inference scheme for the new generation of forward-only techniques, we further analyze our proposed lightweight inference scheme for neural networks trained based on \gls{BP}, MP, OP, and PT.

First, we investigate the distance between the negative data and the positive data for the four datasets considered in this work.  
Here, we calculate the mean values of the negative data and positive data for MNIST, CIFAR-10, CHB-MIT, and MIT-BIH datasets. These mean values are in consecutive layers based on MP. As shown in Fig. \ref{fig:MP_threshold_distance}, the distance between the mean values of the negative and positive data increases as we consider more layers. The figure shows that a higher confidence level is attained as more layers are taken into account.

Then, we investigate the relationship between the Mean Layers and the confidence threshold to show the computational efficiency of our proposed lightweight inference scheme. 
We conduct seven experiments separately with different confidence thresholds for the lightweight inference schemes based on MP, OP, and PT for their default model. At each layer, we extract the \emph{mean} and \emph{std} from validation data and then use the threshold for confidence measurement. The different confidence thresholds range from $mean-3\cdot std$ to $mean+3\cdot std$, \cameraready{where $mean- std$ is the default setting}. As shown in Fig \ref{fig:different_threshold}, if the confidence threshold increases, the mean number of layers used increases accordingly. The reason is that a higher confidence threshold means that fewer test samples in the inference process are regarded as confident and need to be passed to the next layers. Hence, our lightweight inference scheme provides the possibility to achieve different levels of computational efficiency for different confidence thresholds.

We also investigated the probability of the number of layers used for Light-MP, Light-OP, and Light-PT.
Towards this, we extract the probability of each layer used for all the test data based on confidence. We refer to the same experimental settings in Table \ref{tab:goodness_1}. Fig \ref{fig:different_layer_probability} shows the probability of the number of layers used in the proposed schemes. We use the lightest color to represent one layer and the darkest color to represent six layers. For the MNIST dataset, more than 77\% of the test samples are confident just at the first layer, for Light-MP, Light-OP, and Light-PT considering different number of layers. For CIFAR-10 datasets, more than \baichuanchange{50}\% of the test samples are confident at the first layer for Light-MP, Light-OP, \cameraready{and Light-PT} with different numbers of layers. For CHB-MIT and MIT-BIH datasets, respectively, more than \baichuanchange{32}\% and  69\%  of the test samples are confident at the first layer for our proposed lightweight inference scheme based on MP, OP, and PT. Fig \ref{fig:different_layer_probability} shows that by considering confidence, our lightweight inference scheme decreases the mean number of layers used in inference.

Finally, let us apply the proposed lightweight inference scheme for a network trained based on \gls{BP} and compare the error against our lightweight inference scheme for networks trained based on FF. Although early-exit strategies at the intermediary layers has been studied in deep neural networks trained based on \gls{BP}, such strategies generally degrade the classification performance, according to our experiments shown in Table \ref{tab:bp}. Our lightweight inference scheme, on the other hand, makes use of the inherent nature of forward-only algorithms, such as FF, that extract relevant features (for the final prediction outcome) early on in deep neural networks, e.g., local energy-based methods. Therefore, our lightweight inference scheme reduces the inference time and complexity of networks trained based on forward-only algorithms, without any major degradation in the classification performance, while such strategies may deteriorate the performance of networks trained using \gls{BP}.

\begin{table}[ht]
\vspace{0.4cm}
\caption{Our Lightweight Inference Scheme for \gls{BP}}\vspace{0.2cm}
\centering
\resizebox{\linewidth}{!}{
\begin{tabular}{c|cc|ccc}
\toprule[2pt]
\multirow{2}{*}{Dataset}             & \multicolumn{2}{c|}{BP\cite{rumelhart1986learning}} & \multicolumn{2}{c}{FF\cite{hinton2022forward}} &  \\ \cline{2-6} 
                                     & BP      & Light-BP     & FF      & Light-FF     &  \\ \midrule[1pt]
MNIST \citep{lecun1998mnist} &    \underline{1.33}\text{\tiny$\pm$0.04}     &    5.21\text{\tiny$\pm$0.69}     &    1.53\text{\tiny$\pm$0.10}     &   \underline{1.02}\text{\tiny$\pm$0.07}           &  \\
CIFAR-10 \citep{krizhevsky2010cifar} &    \underline{43.62}\text{\tiny$\pm$0.33}     &   54.22\text{\tiny$\pm$0.11}           &   47.78\text{\tiny$\pm$0.37}      &        \underline{46.25}\text{\tiny$\pm$0.37}      &  \\
CHB-MIT \citep{shoeb2009application} &    \underline{25.63}\text{\tiny$\pm$0.40}     &      40.69\text{\tiny$\pm$0.76}        &    37.98\text{\tiny$\pm$0.98}     &     \underline{28.23}\text{\tiny$\pm$0.08}         &  \\
MIT-BIH\citep{mark1982annotated}     &   \underline{8.25}\text{\tiny$\pm$0.46}      &   11.55\text{\tiny$\pm$0.09}           &    10.86\text{\tiny$\pm$0.13}     &       \underline{10.54}\text{\tiny$\pm$0.14}    &  \\ \bottomrule[2pt]
\end{tabular}}
\label{tab:bp}
\end{table}

\section{Related Work}
Over the past years, various biologically plausible alternatives have been proposed to address the inherent biologically-implausible nature of back-propagation, e.g., \cite{lillicrap2016random,nokland2016direct, liao2016important,schiess2016somato, scellier2017equilibrium, whittington2017approximation, guerguiev2017towards, sacramento2018dendritic, nokland2019training, meulemans2021credit, millidge2022predictive, hinton2022forward, journe2022hebbian, lee2023exact, dellaferrera2022error, chan2022redunet, guo2017calibration}.
  
In \cite{hinton2022forward}, Hinton proposes one of the recent alternatives for back-propagation, called the Forward-Forward algorithm, which is suitable to be implemented on low-power analog hardware. 
In the Forward-Forward algorithm, the forward and backward passes of back-propagation are replaced by two forward passes. This algorithm addresses, at least partially, four well-known problems in learning with backpropagation, i.e., weight transport \citep{grossberg1987competitive},  non-local weight update \citep{whittington2019theories}, frozen activity \citep{lillicrap2020backpropagation}, and update locking \citep{czarnecki2017understanding,jaderberg2017decoupled}.

In \cite{dellaferrera2022error}, Dellaferrera and Kreiman propose a similar scheme to the Forward-Forward learning, called PEPITA, executing the forward pass twice. In the second forward pass, the input data is modified by the output error of the first forward pass, while the Forward-Forward scheme does not use any feedback. 
In \cite{journe2022hebbian}, Journe et. al also consider two forward passes for the learning procedure and proposes an unsupervised Hebbian-based learning algorithm. 
Moreover, several very recent studies propose to use the modified versions of the Forward-Forward algorithm for the learning procedure, e.g.,
\cite{ororbia2023predictive, zhao2023cascaded, psenka2023representation,baghersalimi2023layer,flugel2023feed}.

Lightweight inference has also been considered in several state-of-the-art studies. Several lightweight inference schemes have been developed both for classical DNNs \citep{park2015big, venkataramani2015scalable,panda2016conditional,georgiev2017low, bolukbasi2017adaptive,huang2023epilepsynet,antoran2020depth} and for classical feature-based machine-learning techniques \citep{Sopic18TBIOCAS, Forooghifar2019MONET, zanetti2021real, forooghifar2021self, huang2023lightweight}. 
For instance, early exit and cascade networks, which incorporate mechanisms for making predictions at intermediate stages of processing, have been proposed to reduce the inference time and energy \cite{teerapittayanon2016branchynet,matsubara2022split, scardapane2020should, laskaridis2021adaptive, sabokrou2017deep}. 
However, lightweight inference for the forward-only techniques has not been addressed to date. In this work, we bridge this gap and develop the first lightweight inference scheme for neural networks trained based on forward-only techniques.

\vspace{-0.5cm}
\section{Conclusions}
In this paper, we propose a lightweight inference scheme based on the new generation of forward-only techniques, measuring the confidence at each layer. This lightweight inference scheme aims for computational efficiency in the inference of these forward-only \gls{DNNs}. We have evaluated our lightweight inference scheme on the Forward-Forward Multi-Pass \citep{hinton2022forward}, One-Pass \citep{hinton2022forward}, and PEPITA \citep{dellaferrera2022error}, based on the MNIST, CIFAR-10, CHB-MIT, and MIT-BIH datasets. The results show that our lightweight inference scheme achieves computational efficiency by decreasing the number of layers used in inference, with a comparable classification error.

Our work aims at developing resource/energy-efficient inference mechanisms for modern AI/ML, towards a new generation of sustainable AI/ML techniques. Moreover, our proposed inference scheme enables the adoption of machine learning techniques by resource-constrained wearable devices and enables real-time and long-term health monitoring, on a personalized basis. As such, our work also contributes to the realization of the ``precision medicine'' paradigm. To demonstrate this, we have considered two health applications to showcase our proposed inference scheme even in the context of real-world medical applications.

\textbf{Limitations.} Despite their recent success, as we also show for two real-world health applications, the state-of-the-art forward-only techniques are only in their infancy and are yet to be developed to be able to tackle the most complex/challenging learning tasks in the domain. As a result, our proposed inference scheme is also limited to the extent the state-of-the-art forward-only training techniques apply.

\textbf{Broader Ethical Impact.} There are no potential ethical impacts and future societal implications/consequences to be highlighted here.

\section*{Acknowledgements}
This research has been partially supported by the Swedish Wallenberg AI, Autonomous Systems and Software Program (WASP), the Swedish Research Council (VR), Swedish Foundation for Strategic Research (SSF), the ELLIIT Strategic Research Environment, and the European Union (EU) Interreg Program. The computations were enabled by resources provided by the National Academic Infrastructure for Supercomputing in Sweden (NAISS) partially funded by the Swedish Research Council through grant agreement no. 2022-06725.



\bibliography{mybibfile}

\end{document}